# Neither "hear" nor "their": Interrogating gender neutrality in robots

**KATIE SEABORN**

*Industrial Engineering & Economics*
*Tokyo Institute of Technology*

**PETER PENNEFATHER**

*gDial, Inc.*

# Neither "Hear" Nor "Their": Interrogating Gender Neutrality in Robots

Katie Seaborn
*Department of Industrial Engineering and Economics*
Tokyo Institute of Technology
Tokyo, Japan
0000-0002-7812-9096

Peter Pennefather
*gDial, Inc.*
Toronto, Canada
0000-0002-2953-8977

*Abstract*—Gender is a social framework through which people organize themselves—and non-human subjects, including robots. Research stretching back decades has found evidence that people tend to gender artificial agents unwittingly, even with the slightest cue of humanlike features in voice, body, role, and other social features. This has led to the notion of gender neutrality in robots: ways in which we can avoid gendering robots in line with human models, as well explorations of extra-human genders. This rapid review critically surveyed the literature to capture the state of art on gender neutrality in robots that interact with people. We present findings on theory, methods, results, and reflexivity. We interrogate the very idea that robot gender/ing can be neutral and explore alternate ways of approaching gender/ing through the design and study of robots interacting with people.

## I. Introduction

Gender is a primary way in which people organize the world, including other people but also non-human agents, from other animals to cartoons to robots. A wealth of work on computer-based artificial agents has shown that people will tend to ascribe gender, often without realizing it, based on the slightest of cues. This idea was captured in the work of Nass, Brave, Moon, Lee, and colleagues [1], [2] on computer voice: the Computers Are Social Actors (CASA) model. This descriptive paradigm shows how people unthinkingly react to artificial agents, like robots, as if they are people. Notably, these reactions tend to align with stereotypes of human gender. A recent review on computer voice in robots and other agents indicates that while *stereotypes* may shift over time, the tendency for people *to stereotype* agents based on perceived gender has not [3]. In short, how we organize each other through the framework of gender also seems to be how we organize artificial agents, including robots.

One implication of this work is the idea that robots and other artificial agents may be gendered in extra-human ways. Most of the work so far has relied on human models of gender. In particular, most have taken a binary approach, with man/male/masculine and woman/female/feminine poles. Yet, alternative models may be more appropriate. Ambiguous, non-binary, fluid, mechanical/robotic, and neutral genders are possibilities with some empirical backing and theoretical grounding [4]–[6]. The extent to which the voices of robots and other agents can be perceived as genderless, under what conditions and by whom, and whether this is an effective and stable approach, remain pressing questions There is yet no consensus on the state of affairs and what is needed to properly explore these concepts.

The notion of gender neutrality has also been characterized as an absence or removal of gender, effectively becoming a way in which to prevent or disrupt gender-based stereotypes and harmful associations through avoidance [6]. If one eliminates all "gender cues," the argument goes, then people will not gender the robot. Thus, any negative side-effects associated with gender/ing will be avoided. Yet, some have argued that artificial agents cannot be genderless due to our propensity to assign gender when any human-like cues are present [5]. Aldebaran-SoftBank's Pepper robot is a prime example. The designers' stated aim was to create a gender-neutral robot. However, the official materials gender Pepper using masculine pronouns[1]. Indeed, a critical analysis of how researchers have gendered Pepper indicates that, wittingly or not, Pepper was not always taken for granted as gender neutral, by participants or the researchers themselves [7]. How gender neutrality has been approached as a tool against gendering and to what extent it has been effective in reducing or eliminating undesired, stereotyped, and negative reactions remains unknown.

This state of affairs has led to the creation of a new imitative: a *living review* on gender neutrality in human-robot interaction (HRI) research [8]. Living reviews, unlike other kinds of reviews, are not static. Instead, they are regularly updated as new research is produced [9], [10]. For the first cycle in this living review project, we conducted a rapid review using the associated open access dataset[2]. Our overarching research question was: *What is the state of research on gender neutrality in HRI?* We generated a series of sub-questions and hypotheses to map out the situation from theoretical, knowledge, methodological, and reflexive perspectives. This multi-pronged approach to inquiry allowed us to make several contributions. First, a critical, reflexive, and evidence-based report on the state of affairs. Second, a nuanced view on the gender question, showing support for gender neutrality, gender ambiguity, and CASA-compliant gendering. Third, pathways for future research. This work is expected to help direct ongoing efforts in exploring a more creative range of robot designs as well as robots for the social good. It joins the call against assuming neutrality, arguing for more critical and socially aware approaches to the design of robots and other artificial agents that interact with people.

---

[1] https://www.softbankrobotics.com/emea/en/pepper

[2] https://osf.io/v6fwg

## II. METHODS

We carried out a rapid review of the literature according to Cochrane's guidelines [11] and in line with the living review [8]. We used[3] the PRISMA approach [12]. Rapid reviews are "a type of knowledge synthesis in which [systematic review] methods are streamlined and processes are accelerated to complete the review more quickly" [11, p. 14]. Typically, systematic reviews take years to complete and involve conducting a scoping review first. Given the recent and ongoing influx of work on gender neutrality in research on robots and other computer agents, we determined to rapidly capture the state of affairs before we get ahead of ourselves. Details of the protocol are available in [8]; we provide only the specifics for this cycle of the living review in this paper. This protocol was registered before queries were run on OSF[4] on September 10th, 2021. Our PRISMA flow chart can be found on OSF[5].

### A. Research Questions and Hypotheses

We asked a set of research questions (RQs) and hypotheses related to three major aspects of research practice—theory, methods, and findings—plus reflexivity as a meta-level factor. These allow capturing of an overview of the kinds of knowledge and praxis available. With these, we can better understand what we know and do not know so far about gender neutral robots.

*1) Theory*

We considered theoretical foundations as a grounding of the research, asking RQ1-1: *What theories of gender neutrality have been used?* Given that this is a novel area of study where theories may not yet exist, we also considered other ways in which researchers may justify their motivation and contribution, asking RQ1-2: *What reasons were given to explore gender neutrality in robots?* The design of the robot should be based in theory and reasoning, so we asked RQ1-3: *What features of robot design were used to invoke gender neutrality?*

*2) Methods*

Methodological integrity relates to how gender neutrality was explored, so we asked RQ2: *How has gender neutrality been evaluated?* We considered the use of manipulation checks, gender response options used, and, given the results of recent review work [3], evaluation of the voice and body of the robot.

*3) Findings*

Consensus is needed to steer the research, so we asked RQ3-1: *Have gender neutral robots been achieved?* We optimistically expected so (RQ2-1-H1). We also needed to account for other gender/ing possibilities, so we asked RQ3-2: *What alternatives have been explored or captured?* Given recent review work on agent voice [3], we asked RQ3-3: *Is there a relationship between voice and body?* If bodily gender cues exist in the robot, they should match participant gendering of the robot voice, even if the voice is neutral, and vice versa (RQ3-3-H1). Age was also highlighted as a potential characteristic of gender neutrality, so we asked RQ3-4: *What intersections are there with age?*[6] Also, voices perceived as childlike tended to not be gendered, and so we take that position, as well (RQ3-4-H1).

*4) Reflexivity*

Finally, we turn to the meta factor: reflexivity. As noted above, a variety of gender models and orientations exist, and individuals may not always agree on the gender (if any) of a given robot. Indeed, the critical review of Pepper urges us to consider this explicitly [8]. We therefore asked RQ4: *Is there a difference between researcher expectations and participants' ascriptions of gender neutrality in robots?* If gender neutrality is not a given, then researcher expectations will tend not to match participants' gendering of the robot (RQ4-H1).

### B. Eligibility Criteria

We used the same eligibility criteria as in [8].

### C. Information Sources

Four databases representing general and engineering topics were searched: Scopus, Web of Science, IEEE Xplore, and ACM Digital Library. Searches were conducted on September 15th and September 16th, 2021. Papers known by the authors or suggested by experts were manually included. Bibliographies of included papers were searched for missed papers.

### D. Search Terms

We used the same search terms as in [8]. The queries for each database can be found on OSF[4].

### E. Study Selection

The title and abstracts of potential papers were dual-screened independently. Conflicts were resolved through discussion. Each assessed 20% and the first author assessed the rest. The full text of each paper was then independently assessed by the first author, and the second author double-checked excluded papers. All included and excluded papers can be found on OSF[7].

### F. Data Extraction

Each author extracted data for roughly 50% of the papers. Each author then checked the extractions of the other for completeness. Data was extracted into a shared Google Sheet.

### G. Risk of Bias and Quality Assessment

As per [8], we used the Quality Assessment for Diverse Studies (QuADS) tool [13]. The first author independently evaluated the articles included at the full text stage. The second author checked the ratings. There were no disagreements.

### H. Data Analysis/Synthesis

We used descriptive and inferential statistics for the quantitative data. For the qualitative data, we used thematic analysis [14]. Two raters analyzed the data. Initial themes were created by the first author and then applied by two raters independently. Given the low n, inter-rater reliability was based on percentage agreement. Authors discussed themes for which agreement was almost met (above 50% but under 80%). All themes achieved at least 89% agreement after two rounds.

---

[3] With small changes, e.g., no structured abstract, PICOS, etc.
[4] https://osf.io/gcesd
[5] https://osf.io/7mru3
[6] We deviated from our protocol by not considering tonality, as there was not enough data on it. We also added RQ1-3.
[7] https://osf.io/v6fwg/files

## III. RESULTS

A total of 18 articles covering 19 studies were selected from an initial set of 551 records. Details are available on OSF[8]. There was a total of 1559 participants, with 769 men, 791 women, and 8 of another gender. The average age was 27.6 (SD=10.4, MD=29.8, IQR=17), with ages ranging from five years to 73 years of age. The average quality score was 18.7 (SD=4.7, MD=18, IQR=11), with a min of 7 and a max of 27 (range=0-39). We now present the results per RQ and hypothesis.

### A. What theories of gender neutrality have been used?

Two papers (11%) used a theory: uncanny reactions to robots, used by Otterbacher & Talias [15], and categorization ambiguity theory, used by Paeztel et al. [16]. Both relate to the perceived gender of the robot, i.e., as having a gender or not.

### B. What reasons were given to explore gender neutrality?

Seven of 19 studies (33%) did not include a reason, while 8 had an explicit reason and in 4 the reason was implied. Reasons were categorized as exploring (11 or 58%), controlling (3 or 16%), and avoiding (5 or 26%). 6 papers [15], [17]–[22] aimed to explore participant gendering of robots, one [19] aimed to explore researcher gendering, and 4 [17], [20], [21], [23] aimed to explore the possibility of gender neutrality in robots. 3 studies [15], [18], [20] used gender neutrality as a way to focus on a specific feature of the robot. In 4 studies [21]–[24], gender neutrality was used to avoid participant gendering of robots, and in one [17] it was used to avoid researcher gendering of robots.

### C. What features were used to invoke gender neutrality?

7 papers (39%) did not explain how gender neutrality was designed for. 5 papers relied on the name, 4 on the voice, 3 on the hair, 2 on the colour of the body, and one each for face, lack of body or voice, and incongruent gender cues.

### D. How has gender neutrality been evaluated?

Seven papers (39%) conducted a manipulation check of gender. Of these, most were of the perceived gender of the robot/s. One related to the robot's occupational role [17]. Others explored perceptions of robot gender [15], [25] or gendering through stereotyped task assignment [26] as the main research method. 10 papers (56%) did not evaluate gender neutrality directly, rather taking for granted that it was achieved. 15 papers (83%) provided predefined gender response options, while two assigned participant gendering without participant involvement [21], [24], and one assumed perceived gender neutrality [27]. Five provided gender binary options. Gender neutral options included "neutral," "neither," "androgynous," "gender-neutral," "no cues," "mechanical," and "humanlike" with no mention of gender. 9 papers (50%) assessed voice, 14 (78%) assessed body, 6 (33%) assessed both, and two assessed the context.

### E. Have gender neutral robots been achieved?

This question was difficult to answer because participant gendering results were not reported or not described in enough detail. Also, given the variety of research conducted, generalizing quantitatively was not possible.

---

[8] https://osf.io/vhtw3

Five studies [15], [22], [23], [25], [28] provided evidence that gender neutrality was achieved. Otterbacher et al. [15] showed that robots without gender cues did not evoke uncanny responses. Chita-Tegmark et al. [23] found that gender neutral names were equally likely to be associated with low or high perceived emotional intelligence. Blancas et al. [25] found that children drew genderless rather than binary gendered robots. Strait et al. [28] found that the gender neutral robot elicited fewer negative comments. Warta [22] found that the gender neutral robot was more likeable. Note that none of these studies reported on a manipulation check, so it is unknown whether the results are due to the robot's gender neutrality or other factors, such as the measures or tasks being perceived as gender-neutral.

Eight studies provided mixed evidence. For instance, Bryant et al. [17] found that 33% of participants classified the gender neutral robot as male. In Jewell et al. [24], half of the children gendered the robot male or female. Other research provided indirect results. For example, the "moderately humanlike, no gender cues" robot in Hover et al. [19] was perceived as a threat to human jobs and safety compared to binary gendered alternatives, suggesting that the robot was perhaps gender neutral, or at least not masculine or feminine.

Three studies [20], [29], [30] found evidence against gender neutrality. For instance, Rogers et al. [30] found that 15 out of 15 participants (30%) assigned a binary gender to the robot. Three studies [18], [31], [32] did not present findings that could be related to the gender neutrality of the robot. Given these mixed and unclear results, we can neither accept nor reject the hypothesis that gender neutral robots are achievable.

### F. What alternatives have been explored or captured?

Gender neutrality has been framed as the absence of gender cues or the use of cues not associated with a particular gender. One potential alternative situated within the binary of model of gender is an equal combination of masculine and feminine gender cues. Paetzel et al. [16] combined differently gendered robot faces and voices, finding that a male face with a feminine voice garnered the largest portion of neutral assignments. Another option is robot name. The results of the study by Chita-Tegmark et al. [23] are described above. Tani et al. [21] found that non-gender neutral names strongly correlated with the gender identity of the assigner, which suggests that robots with gender neutral names may not elicit gendered reactions. Another alternative is the context in which the robot is placed. Bryant et al. [17], for example, found that certain roles were not clearly gendered: tour guide (48%) and news anchor (42%).

### G. Is there a relationship between voice and body?

Evidence was inconclusive due to reporting and methodological limitations. You and Lin [27], for instance, did not ask participants to evaluate the gender of the voice. Steinhaeusser et al. [32] assumed that the body was neutral when evaluating the voice. Sandygulova and O'Hare [20] assumed that the body was gender neutral, but children's ascriptions of gender suggest otherwise. McGinn and Torre [31] speculated that Poli's round head influenced gender neutral ascriptions of its voice, but did not assess this possibility. Paetzel et al. [16] explicitly evaluated different gender cues in voice and body, finding that matched voice and body was most congruent, even while combining male faces with female voices led to more gender neutral ratings. Given the situation, we are not confident in deciding on whether to accept or reject the hypothesis that voice or body being gendered can override gender neutrality.

*H. What intersections are there with age?*

Age can be considered with respect to participant age or perceived age of the robot. For the former, evidence is mixed, with some research indicating that children tend to perceive gender neutrality in the absence of cues [24], [25] or incongruent cues [16]. In Sandygulova and O'Hare [20], younger children compared to older children tended to assign a feminine-ascribed robot as masculine. However, these results may have been influenced by giving children set gender options in line with the gender binary and/or the researchers ascribing gender during analysis in line with the gender binary.

In terms of robot age, Ladwig and Ferstl [26] found that a "child pattern" in the design of robots was associated with gender neutral associations. One study is not sufficient to answer the hypothesis, but it does provide a foundation upon which to explore the intersection of "young" age and gender neutrality.

*I. Is there a difference between researcher expectations and participants' ascriptions of gender neutrality in robots?*

In most cases (12 or 67%), there was not enough information to compare. For researcher ascription of robot gender, 11 papers (61%) suggested neutral gendering based on pronoun usage ("itself") with respect to the robot/s. 6 (33%) cases avoided pronouns. One [31] used gender neutral pronouns for Pepper and feminine pronouns for Poli. Using this data, we can see evidence of a general mismatch. For example, participants in the study by McGinn and Torre [31] gendered the robots exactly opposite to expectations. As such, we can tentatively accept the hypothesis that participant gendering tends not to match that of researchers.

## IV. DISCUSSION

Gender neutrality is a budding subject of study in HRI, and this rapid review clearly shows its infancy. Synthesis and clear conclusions were hard to come by. Yet, we were able to generate a description of the state of affairs, initial evidence, gaps and weakness, and trajectories for future research.

Gender neutrality has been framed as a means to explore, control, or avoid participant and researcher gendering of robots. Yet, it is severely undermotivated and undertheorized, with only two papers citing theory. Given conceptual work, e.g., [4], [33], this shows a massive disconnect between theorizers and empiricists. Empirical work should aim to situate itself against theory, developing research questions and hypotheses based on theories and concepts within and beyond HRI. Work in gender and feminist studies may be translated and extended to HRI contexts as well. Less than half conducted manipulation checks, but these should be a standard of practice to ensure neutrality was achieved and to what degree.

A severe mismatch was found between participant gendering and researcher expectations. This should be considered in light of a significant number of papers not providing justification for the deemed "gender neutral" design of the robot. This must always be done with sound reasoning if not theory or empirical backing, ideally in the system design section. At the same time, researchers tended to prescribe gender through the instruments of study, particularly constrained and limited response options and data analysis. Spiel et al. [34] and Seaborn and Frank [7] offer gender-expansive frameworks that can be used. Defining the how of establishing gender neutral designs and providing ways of capturing perceptions of gender expansively are needed.

Gender neutrality may often be instantiated through the design of robot voice and body. Indeed, one alternative is a form of gender ambiguity or fluidity through combining feminine and masculine characteristics, assuming the gender binary. The research so far also indicates a few alternatives worthy of exploration. Robot name (and possibly the act of naming as a manipulation check) and context, including the scenario of the HRI experience and the role of the robot, are promising options. Additionally, perceived age of the robot may be linked to neutrality. There is much low-hanging fruit to nibble on.

Against expectations, little could be determined about the relationship between robot voice and body. Yet, phenomena like the "vocaloid shift" [3] are drawing attention to the dynamic and mutual influence of human gender models and shifts in the gendering of vocal agents. A masculine bias can be seen in our findings, which is not unexpected for robots [35]. Still, the proliferation of "female" voice assistants, notably Amazon's Alexa and Apple's Siri, has indicated a transition in favour of feminine vocals [3]. We now need to explore the influence of neutrality, and potentially other, non-binary genders, in the mix.

*A. Limitations*

Rapid reviews are limited by their rapid nature. We did not search all databases, and we did not review all reference lists of included papers for manual additions, so it is possible we missed some articles. However, these will be resolved in the next cycle.

## V. CONCLUSION

Gender neutrality in robots is an exciting and nascent subject of study within HRI research and practice, bringing together the fields of robotics and gender studies. Still, as this rapid review shows, much more work is needed, and in more rigorous ways. Gender neutrality is not a given, but it has potential. We should continue to explore it as a design feature, in creative endeavours, as a control in research, and in initiatives for the social good.

ACKNOWLEDGMENT

We thank reviewers of our open framework paper for their critical and insightful feedback, which also improved this paper.